\documentclass[12pt]{article}
\usepackage{amsmath}
\usepackage{amsfonts}
\usepackage{amssymb}
\usepackage{amsthm}
\usepackage{amsbsy}

\usepackage{bm}
\usepackage{nicefrac}
\usepackage{microtype}
\usepackage{lineno}
\usepackage{graphicx}
\usepackage{epsfig}
\usepackage{float}
\usepackage{subfigure}

\usepackage{enumerate}
\usepackage{enumitem}

\usepackage{algorithm}
\usepackage{algpseudocode}

\usepackage{array}
\usepackage{multirow}
\usepackage{makecell}
\usepackage{booktabs}
\usepackage{diagbox}

\usepackage{natbib}
\usepackage{doi}

\usepackage{url}
\usepackage{hyperref}

\usepackage{xcolor}
\usepackage{color}

\usepackage{ulem}
\usepackage{courier}

\usepackage{rotating}
\usepackage{pdflscape}

\setcitestyle{maxcitenames=3, mincitenames=1}

\newcommand{\blind}{1}

  \renewenvironment{thebibliography}[1]{
    \begin{oldthebibliography}{#1}
      \setlength{\parskip}{0ex}
      \setlength{\itemsep}{0.ex}}
  {\end{oldthebibliography}}

\newcommand{\myvec}[1]%
{\stackrel{\raisebox{-2pt}[0pt][0pt]{\tiny$\rightharpoonup$}}{#1}}



\makeatletter
\newcommand*{\centerfloat}{%
  \parindent \z@
  \leftskip \z@ \@plus 1fil \@minus \marginparwidth
  \rightskip \leftskip
  \parfillskip \z@skip}
\makeatother

\newtheorem{theorem}{Theorem}[section]
\newtheorem{proposition}[theorem]{Proposition}

\usepackage{enumitem}

\newtheorem{definition}[theorem]{Definition}

\newtheorem{assumption}{Assumption}[section]

\begin{document}

\def\spacingset#1{\renewcommand{\baselinestretch}%
{#1}\small\normalsize} \spacingset{1}


\if1\blind
{
  \title{\bf  Efficient Online LLM Watermark Detection via Rao–Blackwellized E-Processes}
 \author{
    Lu Luo$^{1}$,
    Dandan Mo$^{1}$,
    Chengdong Xu$^{2}$,
    Ting Li$^{2}$,\\
    Jinhan Xie$^{1}$,
    Huiqiong Li$^{1}$,
     Niansheng Tang$^{1}$,\\
    \\
    $^{1}$Yunnan Key Laboratory of Statistical Modeling
and Data Analysis,\\ Yunnan University, Kunming, China\\
        $^{2}$School of Statistics and Management, \\Shanghai University of Finance and Economics, Shanghai, China\\
}
\date{}
\maketitle

\bigskip
\begin{abstract}
As large language models (LLMs) are increasingly deployed, reliable and efficient mechanisms for distinguishing AI-generated text from human-written content have become essential. Statistical watermarking has emerged as a promising solution, yet most existing methods are typically fixed-horizon procedures,  precluding valid early stopping in streaming generation.  In this paper, we develop an efficient online watermark detection framework with anytime-valid inference based on Rao--Blackwellized e-processes, enabling recursive token-level evidence updates without storing the full history. In particular, we instantiate the framework for the Gumbel-max watermark and reduce the original token-level dependence testing problem to a pivot-induced sequential testing problem with an explicit null distribution. Theoretically, we prove anytime-valid Type I error control under arbitrary optional stopping and establish positive asymptotic log-growth under watermarking, implying consistency of the proposed stopping rules.  Simulations and experiments on real LLM-generated text demonstrate efficient online detection with rigorous anytime-valid guarantees.
\end{abstract}

\noindent%
{\it Keywords: Large Language Models; Anytime-valid; E-values and E-processes; Watermark Detection; ; Rao–Blackwellization} 
\vfill

\newpage
\spacingset{1.9} 

\section{Introduction}\label{sec:intro}
The rapid advancement of large language models (LLMs), exemplified by GPT-4 \citep{achiam2023gpt}, LLaMA \citep{grattafiori2024llama}, and DeepSeek \citep{liu2024deepseek}, has substantially expanded the capabilities of text generation and facilitated the widespread deployment of LLM-based systems, including conversational interaction, summarization, and knowledge-intensive reasoning \citep{brown2020language,bubeck2023sparks,touvron2023llama,grattafiori2024llama}. Yet the same progress that underpin the practical utility of LLMs has also made the provenance of digital text increasingly difficult to ascertain. High-fidelity synthetic text can now be generated and disseminated at scale, creating opportunities for misuse in misinformation campaigns, plagiarism, automated impersonation, and other forms of content abuse \citep{stokel2022ai,kasneci2023chatgpt,milano2023large}. As LLM-generated outputs become progressively more difficult to distinguish from human-authored writing, reliable provenance attribution has emerged as a central challenge spanning statistics, machine learning, and trustworthy AI. This has motivated a rapidly growing literature on principled methods for detecting AI-generated content \citep{jawahar2020automatic,mitchell2023detectgpt,su2023detectllm}.

Broadly speaking, existing detection strategies fall into three main paradigms: machine learning (ML)-based detectors, statistics-based detectors, and watermarking-based detectors. ML-based methods train discriminative classifiers on corpora of human-written and LLM-generated text \citep{guo2024biscope,mao2024raidar}. Statistics-based methods, including DetectGPT \citep{mitchell2023detectgpt} and Fast-DetectGPT \citep{bao2023fast}, instead exploit distributional signatures of generated text, such as token log-probabilities, rank statistics, probability curvature \citep{zhou2025adadetectgpt,zhou2026detecting}. Although these post-hoc approaches have demonstrated empirical effectiveness, their reliability depends heavily on the persistence of model-specific artifacts. As generative models improve, as deployment distributions shift, and as generated text undergoes paraphrasing, editing, or other post-generation perturbations, such artifacts may become attenuated or disappear altogether \citep{li2025statistical}. These limitations motivate watermarking-based methods, which offer a white-box strategy mechanism \citep{tang2024science}: rather than inferring provenance solely from incidental artifacts of generation, watermarking deliberately embeds statistically detectable signals into the LLM decoding processes.

A statistical watermarking scheme implements this principle by modifying the decoding rule of the language model. Let $\mathcal V$ denote the token vocabulary, with $|\mathcal V|=N\ge 2$. At generation step $t$, an unwatermarking language model produces a next-token prediction (NTP) distribution $P_t \in \Delta(\mathcal V)$ and samples a token $W_t \sim P_t$. In contrast, a watermarking model generates the token through a key-dependent decoder,
$W_t = S(P_t,\zeta_t),$
where $\zeta_t$ is a pseudo-random variable (or vector) determined by the previous context and a secret key, and can therefore be reconstructed by an authorized detector. The decoder $S$ is designed to preserve the generation quality while inducing a statistically detectable dependence between $W_t$ and  $\zeta_t$. This dependence-based formulation encompasses representative watermarking mechanisms such as the Gumbel-max watermark and the green-red list watermark \citep{aaronson2023watermarking,kirchenbauer2023watermark}, and it naturally reduces watermark detection to a hypothesis-testing problem: under the null hypothesis, the observed tokens are independent of the reconstructed seeds, whereas under the alternative hypothesis, the watermarking mechanism induces dependence between them \citep{hu2023unbiased,piet2025markmywords}. 

Nevertheless, these developments are primarily tailored to fixed-horizon regimes, in which the detector evaluates a completed text or a prespecified number of tokens. This paradigm is poorly aligned with practical LLM deployments \citep{yao2022react,nakano2021webgpt,xi2025rise}, where outputs are generated sequentially and reliable decisions may be needed before the full response is available. Naively monitoring fixed-sample $p$-values over time can inflate false-positive rates and compromise Type I error control \citep{grunwald2020safe,howard2021time,johari2022always}, while deferring detection to a predetermined horizon may introduce avoidable latency and computational overhead. This tradeoff motivates an anytime-valid sequential watermark detection framework that enables token-by-token evidence accumulation while preserving Type I error control under arbitrary data-dependent stopping rules.

Against this backdrop, recent work has begun to recast the detection of LLM-generated text as a sequential testing problem. \citet{chen2024online} studied online detection of general LLM-generated text through betting-based sequential tests, but their method is not designed for watermark verification and depends on external human reference data as well as an auxiliary scoring function. For watermark detection, \citet{su2026online} developed an e-process framework based on sequential dependence testing between generated tokens and reconstructed pseudo-random variables. Although their approach provides anytime-valid guarantees, its adaptive constructions are inherently history-dependent, i.e., calibrators or weights must be updated from the accumulated sequence of watermark-related statistics, which may limit one-pass, memory-efficient implementations in fully streaming settings. In a related direction, \citet{huang2026towards} proposed an anchored e-value framework for sequential watermark detection, whose performance depends critically on the choice of an anchor distribution used to approximate the target model. Despite these advances, existing approaches do not yet simultaneously provide a fully online and anytime-valid watermark detection procedure that avoids full-history dependence and does not rely on a carefully chosen anchor distribution. This gap motivates the following question:
\begin{quote}
Can one design an \underline{anytime-valid} watermark detector that accumulates evidence \underline{token by token} without retaining the \underline{full history} of watermark-related statistics?
\end{quote}
We answer this question by developing an e-process-based framework for online watermark detection that is both anytime-valid and amenable to one-pass implementation. 
We formulate watermark detection as a sequential test of the dependence induced by watermarking between generated tokens and reconstructed pseudo-random variables, with the Gumbel-max watermark serving as a representative example for instantiating our proposed framework. The key technical step is a pivot-induced reduction motivated by the Rao--Blackwellization principle for e-processes \citep{de2025rao}. This reduction recasts the original watermark testing problem into a tractable reduced form with an explicit null distribution, enabling simple token-by-token updates without retaining the full history of watermark-related statistics.  Building on this, we derive a fixed-parameter e-process and further propose the plug-in and mixture e-processes to address the choice of this fixed parameter.
Figure~\ref{fig:overview} provides an overview of the proposed framework, and our main contributions are summarized below.
\begin{figure}[h]
    \centering
    \includegraphics[width=0.8\textwidth]{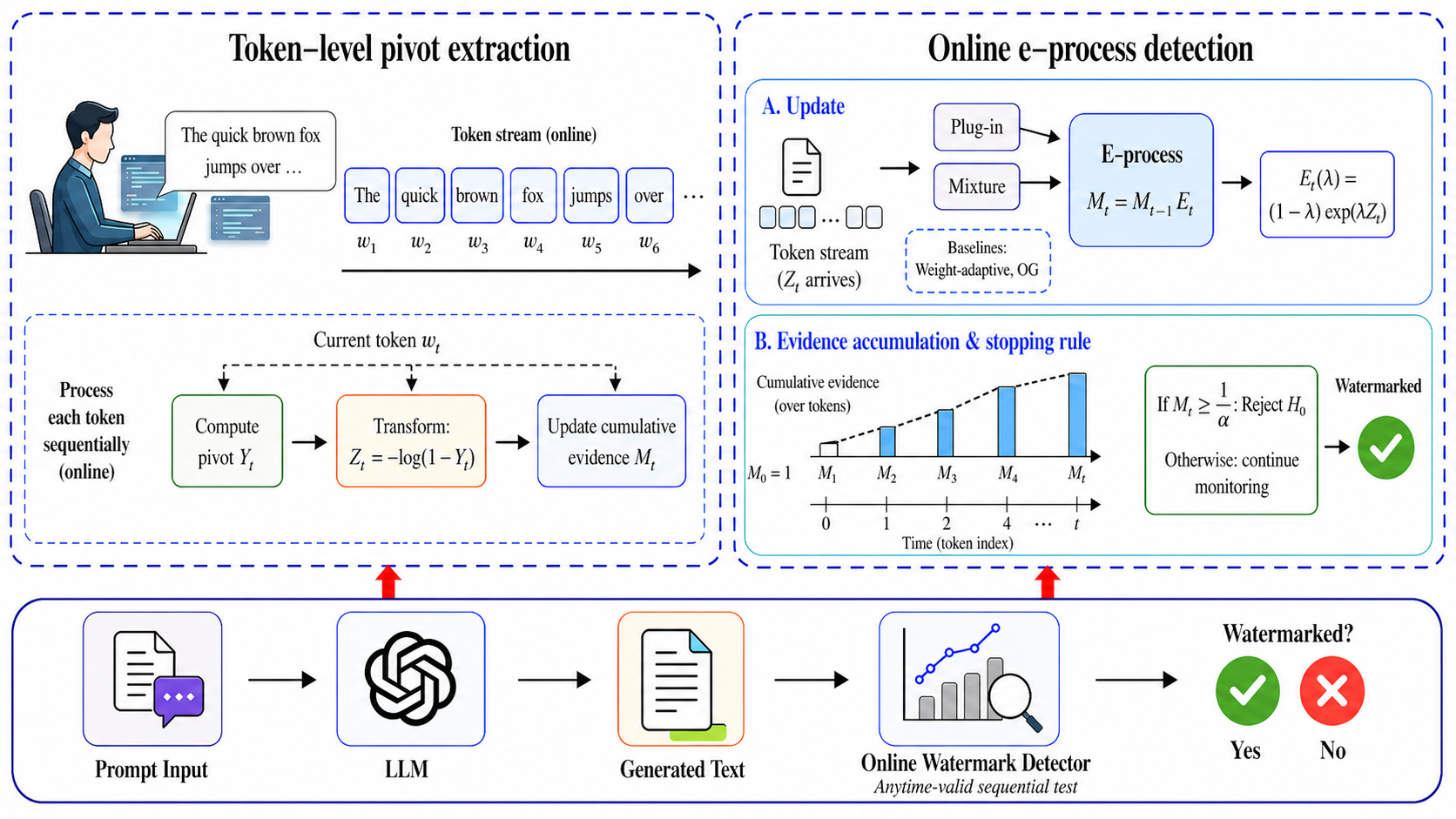}
    \caption{Overview of the proposed online watermark detection framework. The generated text is treated as a token stream. }
    \label{fig:overview}
\end{figure}
\begin{itemize}[leftmargin=*]
    \item \textbf{An online and anytime-valid framework for watermark detection.}
    We develop an anytime-valid framework for online watermark detection based on e-processes. Focusing on the Gumbel-max watermark, we reformulate the problem through a pivot-induced reduced construction under which the null distribution becomes explicit. This leads to a truly online detector without retaining the full history of watermark-related statistics, while preserving validity under anytime stopping.

    \item \textbf{A Rao--Blackwellized reduced formulation.}
    Motivated by the Rao--Blackwellization principle for e-variables and e-processes, we construct a pivot-induced reduced process with an explicit null distribution. Under a surrogate reduced family, the cumulative pivot statistic becomes sufficient,  yielding a formal Rao--Blackwellized interpretation of the proposed e-process.

    \item \textbf{Theoretical guarantees for validity and evidence accumulation.}
    We show that the proposed e-process yields a valid e-process under the reduced null, and hence provides anytime-valid Type I error control under optional stopping. We further establish positive asymptotic log-growth under watermarking, which implies consistency of the corresponding stopping rule. Under the surrogate reduced family, we characterize the oracle value of the optimal e-process in closed form and derive asymptotic guarantees for both the empirical plug-in and mixture extensions.
\end{itemize}

 \section{Preliminaries}\label{sec:preliminaries}

\subsection{E-values and e-processes}
\label{sec:evalues}

To formalize our online watermark detection, we adopt the framework of e-values and e-processes, which enables sequential accumulation of statistical evidence with time-uniform error guarantees \citep{vovk2021values,ramdas2023game,ramdas2025hypothesis}. 

Let $H_0$ denote the null hypothesis, represented by a possibly composite family of distributions $\mathcal P_0$. In a sequential setting, let $\{\mathcal F_t\}_{t\ge 0}$ be the filtration generated by the observations up to time $t$. We begin by recalling the relevant definitions, i.e., e-values, sequential e-values and e-processes.
\begin{definition}[E-value \citep{ramdas2025hypothesis}]
    An e-value $E$ for $\mathcal P_0$ is an $[0,\infty]$-valued random variable satisfying
    $\mathbb E_P[E]\le 1$ for all $P\in\mathcal P_0$ .
\end{definition}

 \begin{definition}[Sequential e-value \citep{ramdas2025hypothesis}]
     A sequence of e-values
$\{E_t\}_{t\ge1}$ for $\mathcal P_0$, adapted to
$\{\mathcal F_t\}_{t\ge0}$, is called sequential if
    $\mathbb E_P\!\left[E_t\mid \mathcal F_{t-1}\right]\le 1$ for all  $t\ge1$  and all $P\in\mathcal P_0$.
 \end{definition}
 
\begin{proposition}
  [E-process \citep{ramdas2025hypothesis}]
Let $\{E_t\}_{t\ge1}$ be a finite (or infinite) sequence of sequential e-values
for $\mathcal P_0$. Define $M_t=\prod_{s=1}^t E_s$
for $ t\ge1$, and  $M_0=1$.
Then $M=\{M_t\}_{t\ge0}$ is an e-process for $\mathcal P_0$.  
\end{proposition}


The resulting anytime-valid guarantee follows from Ville's inequality \citep{ramdas2023game}. For any significance level $\alpha\in(0,1)$,
$\sup_{P\in\mathcal P_0}
    \mathbb{P}\!\left(\sup_{t\ge 0} M_t \ge {1}/{\alpha}\right)
    \le \alpha.$
As a result, the stopping rule
$\tau_\alpha=\inf\left\{t\ge 1: M_t\ge {1}/{\alpha}\right\}$ defines an anytime-valid level-$\alpha$ sequential test. In online watermark detection, this permits continuous monitoring of a growing token stream and valid stopping whenever sufficient evidence has accumulated, without inflating the Type I error probability. 

The subsequent construction is guided by the Rao--Blackwellization principle for e-processes \citep{de2025rao}, which states that conditioning on a sufficient reduced filtration preserves the e-process property and hence maintains anytime-valid inference. This principle allows us to replace a full-data evidence process with a reduced one that removes nuisance randomness and supports simpler online updates. Additional preliminaries on Rao--Blackwellization are provided in Appendix.



\subsection{Problem formulation}
\label{subsec:problem}

We now formulate our online watermark detection problem. At time $t$, the language model conditions on the previously generated text $\bm{w}_{[t-1]}=(w_1,\dots,w_{t-1})$ and produces a NTP distribution $P_t=\mathbb P(W_t=w \mid \bm{w}_{[t-1]})\in\Delta(\mathcal V)$, where $\Delta(\mathcal V)$ denotes the set of probability distributions over $\mathcal V$. In the absence of watermarking, the next token is sampled according to $P_t$, i.e., $W_t\sim P_t$. 

A watermarking mechanism modifies this sampling rule through a secret-key-dependent randomization. Let $K$ denote the watermark key, and let $\zeta_t$ be the pseudo-random variable associated with step $t$, generated from the key and the preceding context. An authorized detector can reconstruct $\zeta_t$ from the observed text and the shared key. The watermarking LLMs generate the next token according to the rule $W_t\sim S( P_t,\zeta_t)$ for some (deterministic) decoder $S$. The purpose of the watermark is to preserve the quality, and ideally the marginal behavior of the base model while inducing a detectable statistical dependence between $W_t$ and $\zeta_t$. This leads to the sequential hypothesis-testing formulation: 
\[
    H_0:~ 
    W_t \perp\!\!\!\perp \zeta_t \mid \mathcal F_{t-1},
    \qquad
    H_1:~ 
    W_t \sim S(P_t,\zeta_t).
\]
Under $H_0$, the pseudo-random variable does not influence token generation; equivalently, conditional on the past, the token is generated independently of $\zeta_t$. Under $H_1$, the watermarking mechanism couples the generated token with the pseudo-random variable via the key-dependent decoding rule.

In our online setting, the detector observes tokens sequentially and updates its evidence after each newly generated token. Our primary goal is to reject $H_0$ as early as possible when watermark-induced dependence is present, while maintaining valid Type I error control under arbitrary data-dependent stopping rules.

\section{Online Watermark Detection}\label{sec:online_watermark}

In this section, we develop a general e-process framework for online watermark detection, with the Gumbel-max watermark serving as a representative instantiation. Our construction first reduces the full sequential experiment to a pivot-induced reduced problem, enabling an anytime-valid detector with token-level online updates. We then introduce a surrogate reduced family that elucidates the Rao--Blackwellized interpretation of the method. Building on this, we further develop plug-in and mixture implementations for practical parameter choices.

\subsection{The proposed token-level pivot framework}
\label{subsec:general_framework}

At each time $t$, the observed watermarking data $(W_t, \zeta_t)$ consist of a token $W_t$ and an associated pseudo-random variable $\zeta_t$. Following the pivot-based formulation of \citet{li2025statistical}, we consider the null hypothesis $H_0$ corresponding to the unwatermarking generation mechanism. Under $H_0$, the token generation satisfies
$
  \mathbb{P}_{H_0}(W_t = w \mid \mathcal{F}_{t-1}, \zeta_t) = P_t(w), ~ w \in \mathcal{V},
$
where $P_t(w)$ denotes the (generally unknown) NTP distribution. Hence, under $H_0$, the pseudo-random signal $\zeta_t$ does not influence token generation.

To test for the presence of watermarking in streaming data, we adopt an e-process framework for sequential inference. A central difficulty is that the underlying distribution $P_t(w)$ is unknown and historical data may not be directly accessible. To overcome this, we construct a token-level pivot 
\begin{equation}
\label{eq:pivot}
  Z_t = T_t(W_t, \zeta_t),
\end{equation}
and base the construction of  e-processes on the transformed sequence $\{Z_t\}_{t \ge 1}$.
We propose to use the reduced filtration $\mathcal{G}_t = \sigma(Z_1, \dots, Z_t) \subseteq \mathcal{F}_t$.
The pivot function $T_t$ is designed such that the conditional distribution of $Z_t$ given $\mathcal{G}_{t-1}$ is fully specified under $H_0$, even though $P_t(w)$ is unknown. Under the alternative hypothesis, the distribution of $Z_t$ deviates systematically from this reference behavior, enabling detection of watermarking effects.

Existing watermark detectors typically operate on the full detection filtration $\mathcal{F}_t$. For example, \citet{christ2024undetectable} proposed a cryptographic watermarking scheme in which detection is performed using the secret key shared with the watermarking generator. Similarly, \citet{dathathri2024scalable} developed SynthID-Text, whose detector aggregates watermark-induced scores computed from the keyed watermarking  mechanism. In contrast, our proposed framework operates with a reduced filtration rather than the full filtration. This construction is in the spirit of Rao--Blackwellization, leading to a more stable and efficient sequential procedure while preserving validity for online testing \citep{de2025rao}.

\subsection{A Gumbel-max watermark example}
The specific form of $T_t$ and the corresponding null reference distribution depend on the watermarking mechanism \citep{aaronson2023watermarking,kirchenbauer2023watermark}. We next present the construction for the Gumbel-max watermark as a representative example. Our method, however, is not tied to this specific design and extends naturally to other watermarking mechanisms.

We now specialize the above framework to the Gumbel-max watermark and derive the corresponding reduced testing problem explicitly. At time $t$, the LLMs produce a NTP distribution $P_t$. Given the watermark key $K$ and the current generation context, we generate a pseudo-random vector $\zeta_t=(U_{t,w})_{w\in\mathcal V}$, where $U_{t,w}\stackrel{\mathrm{i.i.d.}}{\sim}\mathrm{Unif}(0,1)$ for $w\in\mathcal V$. Under watermarking, the $t$-th token is generated by $W_t=\arg\max_{w:P_t(w)>0}\log U_{t,w}/{P_t(w)}.$
For the watermark, a natural pivot is the pseudo-random variable indexed by the token, 
$Y_t=U_{t,W_t}$.
Under the null, the token $W_t$ is sampled from the original $P_t$, yielding the pivotal null distribution $Y_t\mid \mathcal G_{t-1}\sim \mathrm{Unif}(0,1)$, according to the pivot-based analysis of \citet{li2025statistical}. To obtain a more convenient reference distribution for e-process construction, following \citet{fernandez2023three,li2026robust},
we construct the final pivot to be 
\begin{linenomath}
$$
Z_t=-\log(1-Y_t)=-\log(1- U_{t,W_t}), \quad Z_t\mid \mathcal G_{t-1}\sim \mathrm{Exp}(1) \quad \text{under $H_0$}.
$$
\end{linenomath}

Although the pivot-induced reduced testing problem introduced in Section~\ref{subsec:general_framework} has been extensively studied in offline settings \citep{aaronson2023watermarking,li2025statistical}, existing methods are not directly applicable to streaming text data due to their reliance on access to the full dataset or repeated re-computation over accumulated observations.
To handle streaming data without storing the full history, we exploit the explicit null reference distribution of the pivot sequence to construct one-step e-values and the corresponding anytime-valid e-process in the next subsection.

\subsection{E-process construction}
\label{subsec:evalue_construction}


For concreteness, we also consider the commonly used Gumbel-max watermark, under which larger values of $Z_t$ provide stronger evidence against the null. This naturally motivates an exponential scoring rule of the form $e^{\lambda Z_t}$, where $\lambda \in (0,1)$ tunes the sensitivity to large pivot values: larger $\lambda$ places more weight on extreme observations, while smaller $\lambda$ yields a more conservative update.
To ensure validity under the null, we normalize this score by its null moment generating function, following standard exponential supermartingale constructions for anytime-valid inference~\citep{ramdas2023game,ramdas2025hypothesis}. Denote $S_t= \sum_{s=1}^t Z_s$ as the cumulative sum up to $t$, the resulting one-step e-values are then accumulated multiplicatively to form the e-process
\begin{align}
M_t(\lambda)= \prod_{s=1}^t E_s(\lambda)= (1-\lambda)^t e^{\lambda \sum_{s=1}^t Z_s}= (1-\lambda)^t e^{\lambda S_t}.
\label{eq:gumbel_eprocess}
\end{align}
The following theorem shows that this construction defines a valid e-process on the reduced filtration. 

\begin{theorem}
\label{thm:gumbel_validity}
Suppose that under $H_0$, we have $Z_t \mid \mathcal G_{t-1} \sim \mathrm{Exp}(1)$. Then, for every fixed $\lambda\in(0,1)$, the process $\{M_t(\lambda)\}_{t\ge0}$ defined in \eqref{eq:gumbel_eprocess} is an e-process with respect to $\{\mathcal G_t\}_{t\ge0}$ under $H_0$.
\end{theorem}

Theorem~\ref{thm:gumbel_validity} shows that the reduced null distribution alone is enough to produce a valid sequential detector. In particular, for any target level $\alpha\in(0,1)$, the stopping rule 
$$\tau_\alpha=\inf\left\{t\ge1:\ M_t(\lambda)\ge \frac{1}{\alpha}\right\}$$ is anytime-valid. The theorem also leads directly to an online detection procedure for the Gumbel-max watermark. At each time step, the detector reconstructs the pseudo-random variable associated with the emitted token, transforms it into the reduced pivot $Z_t$, and updates the e-process multiplicatively.

\begin{algorithm}[t]
\caption{Online Gumbel-max watermark detection using an e-process}
\label{alg:gumbel_detector}
\begin{algorithmic}[1]
\Require Significance level $\alpha\in(0,1)$, fixed parameter $\lambda\in(0,1)$.
\State Initialize $M_0 \gets 1$
\For{$t=1,2,\dots,T$}
    \State Obtain pivot $Y_t$.
    \State Calculate $Z_t = -\log(1-Y_t)$.
   
    \State Form the one-step e-value $E_t = (1-\lambda)e^{\lambda Z_t}$.
  
    \State Update the e-process $M_t = M_{t-1}\times E_t$.
    \If{$M_t \ge \alpha^{-1}$}
        \State \Return reject $H_0$ at time $t$
    \EndIf
\EndFor
\State \Return no rejection
\end{algorithmic}
\end{algorithm}

The resulting procedure is summarized in Algorithm~\ref{alg:gumbel_detector}, which makes explicit how the proposed test for Gumbel-max watermark detection is implemented in practice. At each step $t$, the detector constructs a one-step e-value based on the emitted token and updates the e-process accordingly. Notably, the procedure only requires maintaining the current value of $M_t$, without accessing or storing the full data history.

The e-process in \eqref{eq:gumbel_eprocess} not only enables efficient online detection, but also admits a natural interpretation within the framework of Rao–Blackwellized e-processes driven by low-dimensional sufficient statistics. This perspective becomes clear by introducing a simple surrogate parametric family built on the null reference distribution.
Specifically, recall that under the null, the pivot follows $\mathrm{Exp}(1)$ with density $q_0(z)=e^{-z}$. We embed this into a one-parameter exponential family
\begin{align}
q_\kappa(z)
= \exp
\{\kappa z - \psi(\kappa)\} q_0(z),
\quad \psi(\kappa) = -\log(1-\kappa), \quad \kappa \in (0,1),
\label{eq:surrogate}
\end{align}
which simplifies to $q_\kappa(z) = (1-\kappa)e^{-(1-\kappa)z}$. Thus, $\kappa=0$ recovers the null, while $\kappa>0$ induces a one-sided deviation under the alternative. Specifically, under $q_\kappa$, the pivot follows $\mathrm{Exp}(1-\kappa)$, so larger $\kappa$ corresponds to a stronger departure from the null. In this sense, $\kappa$ provides a simple and interpretable measure of discrepancy.



\begin{proposition}
\label{prop:surrogate_sufficiency}
For $t\ge1$, consider the surrogate reduced family $\{q_\kappa:\kappa\in(0,1)\}$ in \eqref{eq:surrogate}. Then $S_t=\sum_{s=1}^t Z_s$ is a sufficient statistic for the reduced experiment at time $t$. Moreover, for
every fixed $\lambda=\kappa\in(0,1)$, the constructed e-process in \eqref{eq:gumbel_eprocess} at time $t$ is a function of the sufficient statistic $S_t$.
\end{proposition}

Proposition~\ref{prop:surrogate_sufficiency} shows that the construction is inherently low-dimensional: all relevant information accumulates through the single scalar $S_t$. Consequently, the proposed e-process is not only computationally efficient, but also statistically principled. Under the surrogate model, it coincides with a Rao–Blackwellized e-process, where inference is optimally based on the sufficient statistic.

\subsection{Choosing $\lambda$ in practice}
\label{subsec:lambda_choice}

The fixed-$\lambda$ e-process in Section~\ref{subsec:evalue_construction} is valid for every $\lambda\in(0,1)$, but its performance depends on how well $\lambda$ matches the (unknown) strength of departure from the reduced null. In practice, it is therefore natural to replace a single fixed choice with either an adaptive plug-in rule or a mixture over candidate values. Both approaches preserve anytime-validity and are standard in e-process-based sequential testing \citep{ramdas2022testing,ramdas2023game}.

Suppose the alternative shifts the pivot toward larger values with $\mathbb E_{H_1}[Z_t]=\mu$. The expected one-step log-growth of the fixed-$\lambda$ factor is
\begin{align}
\mathbb E_{H_1}[\log E_t(\lambda)] = \log(1-\lambda) + \lambda \mu.
\label{eq:log_growth}
\end{align}
Maximizing \eqref{eq:log_growth} over $\lambda\in(0,1)$ yields the oracle choice $\lambda^\star(\mu)=1-\mu^{-1}$. Since $\mu$ is unknown, we estimate it online using the empirical mean $\widehat\mu_{t-1} = (t-1)^{-1}\sum_{i=1}^{t-1} Z_i$ for $t\ge2$, updated recursively via $\widehat\mu_t = t^{-1}[(t-1)\widehat\mu_{t-1} + Z_t]$.
This leads to the plug-in e-process
\begin{align}
M_t^{\mathrm{plug}} = \prod_{s=1}^t (1-\lambda_s)e^{\lambda_s Z_s},
\quad
\lambda_s = \mathrm{clip}\left(1 - \frac{1}{\widehat\mu_{s-1}},\lambda_{\min},\lambda_{\max}\right),
\label{eq:plugin_eprocess}
\end{align}
where $0<\lambda_{\min}<\lambda_{\max}<1$ are fixed truncation constants to ensure numerical stability. Since each $\lambda_s$ depends only on past observations, the resulting process remains adapted and preserves validity under the null. 

A complementary strategy is to avoid committing to a single value of $\lambda$ and instead average over a range of candidates. Let $\pi$ be a probability measure on $(0,1)$. The corresponding mixture e-process is $M_t^{\mathrm{mix}}=\int \prod_{s=1}^t (1-\lambda)e^{\lambda Z_s}\,\pi(d\lambda)$.
Because each fixed-$\lambda$ process is an e-process under $H_0$, the mixture construction also preserves null-validity by convexity \citep{ramdas2022testing,ramdas2023game}. In practice, it is convenient to use a
discrete approximation with grid points
$\lambda^{(1)},\dots,\lambda^{(m)}$ and weights $\pi_1,\dots,\pi_m$ which yields
\begin{align}
M_t^{\mathrm{mix}}=\sum_{k=1}^m\pi_k\prod_{s=1}^t(1-\lambda^{(k)})e^{\lambda^{(k)} Z_s},\quad \pi_k\ge0,\ \sum_{k=1}^m \pi_k=1.
\label{eq:mixture}
\end{align}
Appendix states the corresponding validity guarantee for both methods.

\section{Theoretical Guarantees}
\label{sec:theory}

Under the null, we have established validity for controlling the Type I error. We turn to the alternative and analyze the rate at which the proposed e-process accumulates evidence against the null.




\begin{assumption}\label{assump:nondeg}
The probability of generating each token in the NTP is bounded away from  1 for all $t\geq 1$.
\end{assumption}

\begin{assumption}\label{assump:mdep}
    The sequence $(P_t)_{t\ge1}$ is $m$-dependent for some $m\in\mathbb N$.
\end{assumption}

Assumption~\ref{assump:nondeg} excludes nearly degenerate next-token distributions. When $P_t$ is close to a point mass, token generation becomes nearly deterministic, so the watermark-induced perturbation has only limited influence on the realized token and the pivot signal becomes weak \citep{li2025statistical,su2026online}. Assumption~\ref{assump:mdep} imposes a weak dependence condition on the NTP sequence. A closely related $m$-dependence assumption is adopted in \citet{su2026online} as a stylized model of bounded context dependence and is also consistent with context-dependent watermark mechanisms such as \citet{aaronson2023watermarking,kirchenbauer2023watermark}.

\begin{theorem}
\label{thm:growth}
Suppose Assumptions~\ref{assump:nondeg}-\ref{assump:mdep} hold, under the Gumbel-max alternative, there exists $\lambda\in(0,1)$, such that  $\liminf_{T\to\infty} T^{-1} \log M_T(\lambda)>0$ almost surely. Consequently, for every $\alpha\in(0,1)$, the stopping rule $\tau_\alpha=\inf\{T\ge1:\ M_T(\lambda)\ge \alpha^{-1}\}$ can reject the null almost surely.
\end{theorem}

The quantity $\liminf_{T\to\infty} T^{-1} \log M_T(\lambda)$ captures the asymptotic log-growth rate, which governs the power under the alternative \citep{ramdas2023game, su2026online}. Under mild regularity conditions, the reduced e-process $M_T(\lambda)$ achieves strictly positive asymptotic log-growth, implying that the associated anytime-valid stopping rule is consistent. Similar theoretical results are provided in \citet{li2025statistical}, however, they
 consider a fixed-horizon offline setting, which does not permit anytime rejection with valid Type I error control under optional stopping.

Theorem~\ref{thm:growth} further establishes that, under the Gumbel-max reduced alternative, there exists a fixed choice of $\lambda$ that yields positive asymptotic log-growth. We now turn to the surrogate reduced family in \eqref{eq:surrogate}, where both the asymptotic log-growth rate and the corresponding oracle choice of $\lambda$ admit closed-form expressions.

\begin{proposition}
\label{prop:surrogate_growth}
Suppose that under the surrogate reduced family in \eqref{eq:surrogate}, the true data-generating distribution is $q_{\kappa_0}$ for some $\kappa_0\in(0,1)$. Then, the asymptotic log-growth rate $\lim_{T\to\infty} {T}^{-1} \log M_T(\lambda)$ is uniquely maximized at $\lambda^\star=\kappa_0$, and $\lim_{T\to\infty} T^{-1}\log M_T(\lambda^\star)>0$ almost surely. 
\end{proposition}

Proposition~\ref{prop:surrogate_growth} makes the role of $\lambda$ explicit within the surrogate reduced family. In contrast to Theorem~\ref{thm:growth}, which only guarantees the existence of a favorable fixed choice under the actual Gumbel-max reduced alternative, the surrogate model identifies the oracle choice in closed form, i.e. $\lambda^\star=\kappa_0$, and shows that it maximizes the asymptotic log-growth rate within the fixed-$\lambda$ class. This motivates the practical choices of $\lambda$ used in practice; their corresponding theoretical guarantees are given in Appendix.



\section{Experiments}\label{sec:experiment}
In this section, we evaluate the proposed online detectors on both synthetic and real-data-based experiments, and compare them with a range of online and offline benchmarks.
The proposed methods based on plug-in and mixture e-processes are denoted as \textbf{Plug-in} and \textbf{Mixture}, respectively. Online baselines include the \textbf{weight-adaptive} e-process and the \textbf{OG e-process} of \citet{su2026online}.
The
fixed-horizon baselines are denoted by \textbf{Ars}, \textbf{Log}, and
\textbf{Opt}, corresponding to the sum-based watermark statistics with score
functions $-\log(1-y)$ \citep{aaronson2023watermarking}, $\log y$
\citep{fernandez2023three}, and the optimized Gumbel-max score
\citep{li2025statistical}, respectively. We consider the significance level to be $\alpha=0.05$, and 500 replicates.

\subsection{Synthetic experiments}\label{subsec:synthetic}


We evaluate the proposed online detectors in a synthetic setting. The vocabulary size is fixed at $N=1000$ and the horizon at $T=600$. At each time $t$, the NTP distribution $P_t \in \mathbb{R}^{1000}$ is generated from a spike model: draw $\Delta_t \overset{\mathrm{iid}}{\sim} \mathrm{Unif}(0.001,\delta)$, select a spike token $w_t^\star \in \mathcal V$, and assign $P_t(w_t^\star)=1-\Delta_t$, with the remaining mass $\Delta_t$ distributed over other tokens. The pseudo-random watermark vector is given by $\zeta_t=(U_{t,w})_{w\in\mathcal V}$, where $U_{t,w}\overset{\mathrm{iid}}{\sim}\mathrm{Unif}(0,1)$.

Under $H_0$, tokens are sampled from the unwatermarking model $W_t \sim \mathrm{Cat}(P_t)$, where $\mathrm{Cat}(P_t)$ denotes the categorical distribution with probability mass $P_t$. Under $H_1$, tokens are generated via the Gumbel-max watermark decoder
$
W_t=\arg\max_{w: P_t(w)>0} {\log U_{t,w}}/{P_t(w)}.
$
The parameter $\delta$ controls the concentration of $P_t$. Larger $\delta$ increases the typical value of $\Delta_t$, yielding a flatter distribution and more randomness for the Gumbel-max decoder.
Hence, under the alternative, larger $\delta$ makes the watermark easier to detect than smaller $\delta$.

\begin{figure}[h]
    \centering
    \includegraphics[width=\textwidth]{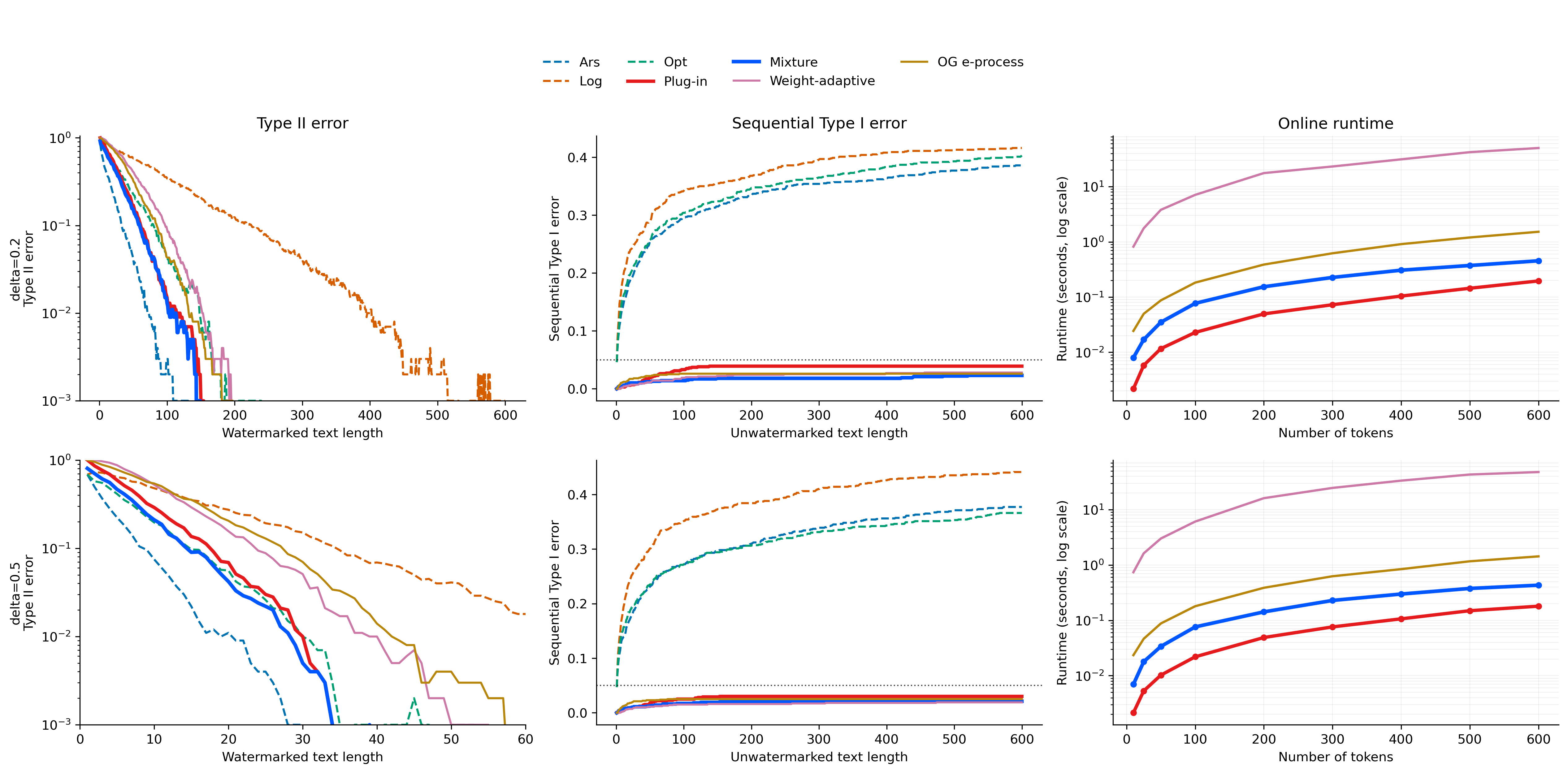}
    \caption{
    Average Type II error (on a log scale), sequential Type I error, and online runtime versus text length/token number on the synthetic data with $\delta=0.2$ (top) and $\delta=0.5$ (bottom).
    }\label{fig:simulation}
\end{figure}

Figure~\ref{fig:simulation} presents the simulation results for $\delta=0.2$ and $\delta=0.5$, reporting the Type II error, sequential Type I error, and online runtime of all online methods. The online methods (solid lines) consistently control the sequential Type I error below the  $\alpha=0.05$, whereas the offline baselines Ars, Log, and Opt (dotted lines) exhibit substantial error inflation in the online setting. Although Ars often achieves the fastest decay of Type II error, this gain comes at the cost of losing sequential validity. 

Among the valid procedures, the proposed plug-in and mixture e-processes show faster reductions in Type II error as the text length increases, with the advantage becoming more pronounced in the stronger-signal setting $\delta=0.5$, while maintaining substantially lower computational cost than the other online procedures. These results suggest that the proposed plug-in and mixture e-processes offer the most favorable trade-off between empirical power, anytime-valid Type I error control, and online efficiency. A complementary endpoint summary is provided in Appendix, showing that the proposed plug-in and mixture e-processes attain power comparable to offline detectors while enabling much earlier stopping.

\subsection{Real-data-based experiments}\label{subsec:real}
We next evaluate the proposed detectors on text generated by real language models under the Gumbel-max watermark. We consider two language models including \texttt{facebook/opt-1.3b} \citep{zhang2022opt} and \texttt{EleutherAI/gpt-neo-1.3b} \citep{black2021gpt}, and generate watermarking text at three sampling temperatures, namely $1.0$, $0.7$, and $0.5$. We use the AG News dataset \citep{zhang2015character} as the prompt source, randomly selecting 500 examples for evaluation.

The methods considered here are identical to those used in Section \ref{subsec:synthetic}, with one additional online baseline included in the real-data setting. In particular, we also evaluate anchored E-watermarking \citep{huang2026towards} under two anchor choices, namely a same-model anchor and a small-model anchor. The former uses the same model as the generator, whereas the latter uses a smaller model from the same family with size reduced to 125M.

\begin{figure}[h]
    \centering
    \includegraphics[width=\textwidth]{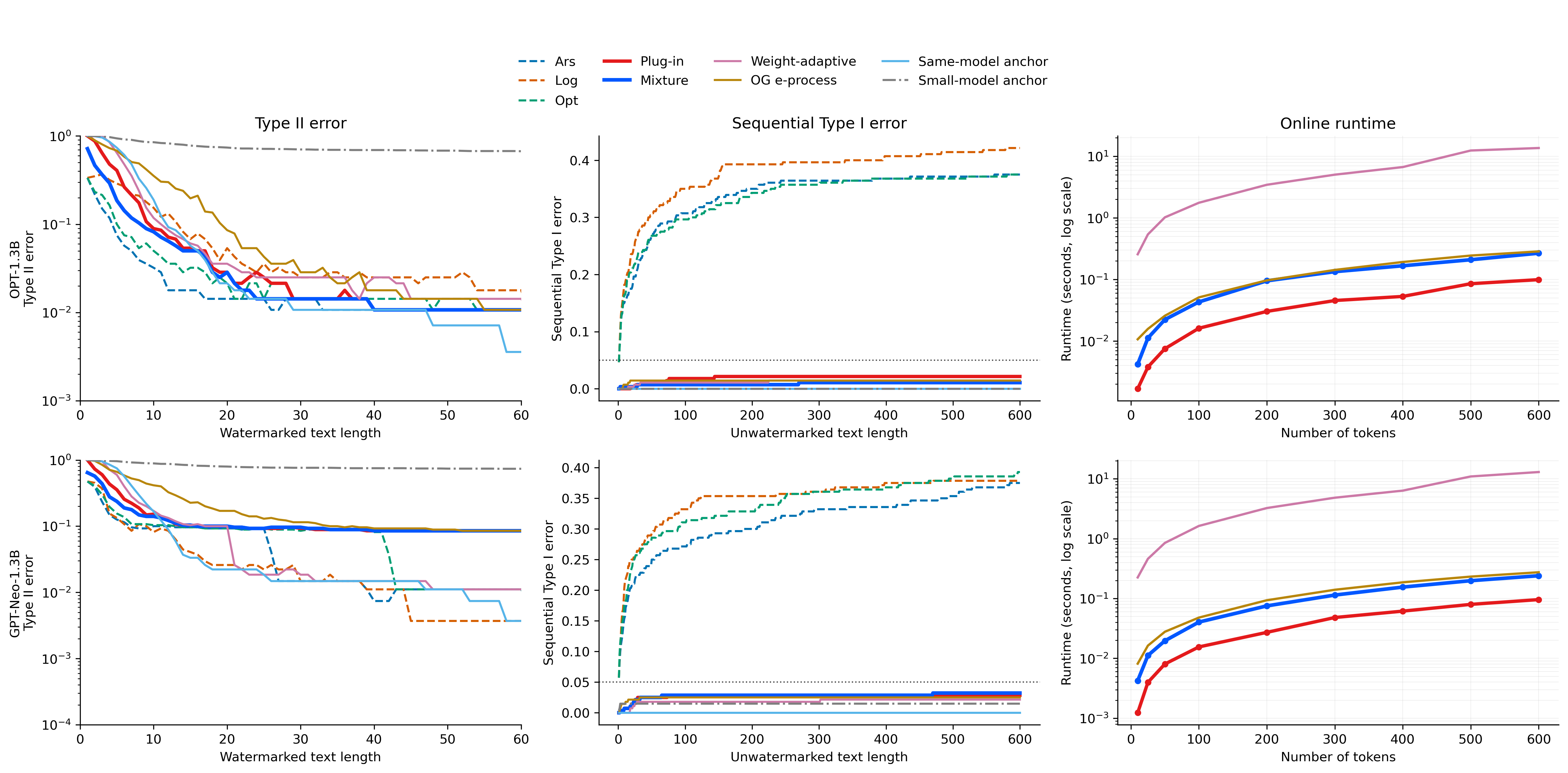}\caption{
    Average Type II error, sequential Type I error, and online runtime versus text length/token number on the real data at temperature $1.0$, with OPT-1.3B (top) and GPT-Neo-1.3B (bottom).
    }
    \label{fig:real_temp1}
\end{figure}

Figure~\ref{fig:real_temp1} reports the real-data results at temperature $1.0$ on OPT-1.3B and GPT-Neo-1.3B. The offline baselines, including Ars, Log and Opt, can achieve low Type II error for sufficiently long watermarking sequences, but their sequential Type I error quickly increases far beyond the nominal level $\alpha=0.05$, making them unsuitable and invalid for online detection. In contrast, the online methods maintain valid sequential Type I error control across both models. Among them, our plug-in and mixture e-processes achieve competitive detection power while being substantially more efficient than the weight-adaptive and OG e-process baselines. The anchor-based comparisons further show that model mismatch can severely weaken detection power, as the small-model anchor exhibits persistently large Type II error, whereas the same-model anchor performs well only when a matched reference model is available. It is important to note, however, that the same-model anchor relies on an anchor distribution obtained by querying the reference LLM, which incurs an additional inference cost that is substantially larger than the post-hoc score-based and e-process computations reported in the runtime panels. Overall, these results demonstrate that our plug-in and mixture e-processes provide a favorable trade-off between empirical power, anytime-valid Type I error control, and online computational efficiency on real LLM-generated text. 

Additional comparisons at temperatures $0.5$ and $0.7$ and endpoint summaries are reported in Appendix, showing that the proposed methods remain competitive across temperatures while preserving the advantage of early stopping.



\setlength{\baselineskip}{0.85\baselineskip}
\bibliographystyle{agsm}
\bibliography{paper-ref}
\end{document}